\documentclass[sigconf]
{acmart}
\AtBeginDocument{%
  }
    
\setcopyright{acmlicensed}
\acmDOI{3774904.3792692}
\copyrightyear{2026}  
\acmYear{2026}
\setcopyright{cc}
\setcctype{by}
\acmConference[WWW '26]{Proceedings of the ACM Web Conference 2026}{April 13--17, 2026}{Dubai, United Arab Emirates}

\acmBooktitle{Proceedings of the ACM Web Conference 2026 (WWW '26), April 13--17, 2026, Dubai, United Arab Emirates}
\acmPrice{}  
\acmDOI{10.1145/3774904.3792692}
\acmISBN{979-8-4007-2307-0/2026/04}
\usepackage{booktabs}
\usepackage{threeparttable}
\usepackage{multirow}
\usepackage{makecell}
\usepackage{algorithm}
\usepackage{subfigure}
\usepackage{xcolor}
\usepackage{listings}
\usepackage{colortbl}
\usepackage[normalem]{ulem}
\useunder{\uline}{\ul}{}
\usepackage{ulem} 
\usepackage{adjustbox}
\usepackage{xcolor}
\usepackage{multirow}
\usepackage{booktabs} 
\usepackage{algorithmicx}  
\usepackage{algpseudocode}  
\usepackage{algorithm}

\usepackage{amsmath}

\usepackage{lineno,hyperref}

\begin{document}

\title{Learning to Evolve: Bayesian-Guided Continual Knowledge Graph Embedding}
\author{Linyu Li}
\affiliation{%
  \institution{Peking University\\School of Computer Science}  
  \orcid{https://orcid.org/0009-0005-8626-0608}
  \city{Beijing}
  \country{China}
}
\email{linyuli@stu.pku.edu.cn}
\author{Zhi Jin}
\authornote{Corresponding author.}
\affiliation{%
  \institution{Peking University\\School of Computer Science}  
    \orcid{https://orcid.org/0000-0003-1087-226X}
  \city{Beijing}
  \country{China}
}
\email{zhijin@pku.edu.cn}

\author{Yuanpeng He}
\affiliation{%
  \institution{Peking University\\School of Computer Science}  
  \orcid{https://orcid.org/0000-0001-9071-2260}
  \city{Beijing}
  \country{China}
}
\email{heyuanpeng@stu.pku.edu.cn}

\author{Dongming Jin}
\affiliation{%
  \institution{Peking University\\School of Computer Science}  
  \orcid{https://orcid.org/0009-0002-8164-6227}
  \city{Beijing}
  \country{China}
}
\email{dmjin@stu.pku.edu.cn}

\author{Yichi Zhang}
\affiliation{%
  \institution{Zhejiang University\\School of Computer Science}  
  \orcid{https://orcid.org/0009-0007-4046-1003}
  \city{Hangzhou}
  \country{China}
}
\email{zhangyichi2022@zju.edu.cn}

\author{Haoran Duan}
\affiliation{%
  \institution{Wuhan University\\School of Cyber Science and Engineering}  
  \orcid{https://orcid.org/0000-0002-2751-2589}
  \city{Wuhan}
  \country{China}
}
\email{haorand@whu.edu.cn}

\author{Xuan Zhang}
\affiliation{%
    \orcid{https://orcid.org/0000-0003-2929-2126}
  \institution{Yunnan University}  
  \city{Kunming}
  \country{China}
}
\email{zhxuan@ynu.edu.cn}

\author{Zhengwei Tao}
\affiliation{%
  \institution{Peking University\\School of Computer Science}  
  \orcid{https://orcid.org/0000-0003-2243-8778}
  \city{Beijing}
  \country{China}
}
\email{tttzw@stu.pku.edu.cn}

\author{Tashi Nyima}
\affiliation{%
  \institution{Tibet University}  
  \orcid{https://orcid.org/0000-0001-9288-6600}
  \city{Lhasa}
  \country{China}
}
\email{nmzx@utibet.edu.cn}


\begin{abstract}
As social media and the World Wide Web become hubs for information dissemination, effectively organizing and understanding the vast amounts of dynamically evolving Web content is crucial. Knowledge graphs (KGs) provide a powerful framework for structuring this information. However, the rapid emergence of new hot topics, user relationships, and events in social media renders traditional static knowledge graph embedding (KGE) models rapidly outdated. Continual Knowledge Graph Embedding (CKGE) aims to address this issue, but existing methods commonly suffer from catastrophic forgetting, whereby older, but still valuable, information is lost when learning new knowledge (such as new memes or trending events). This means the model cannot effectively learn the evolution of the data. We propose a novel CKGE framework, BAKE. Unlike existing methods, BAKE formulates CKGE as a sequential Bayesian inference problem and utilizes the Bayesian posterior update principle as a natural continual learning strategy. This principle is insensitive to data order and provides theoretical guarantees to preserve prior knowledge as much as possible. Specifically, we treat each batch of new data as a Bayesian update to the model's prior. By maintaining the posterior distribution, the model effectively preserves earlier knowledge even as it evolves over multiple snapshots. Furthermore, to constrain the evolution of knowledge across snapshots, we introduce a continual clustering method that maintains the compact cluster structure of entity embeddings through a regularization term, ensuring semantic consistency while allowing controlled adaptation to new knowledge. We conduct extensive experiments on multiple CKGE benchmarks, which demonstrate that BAKE achieves the top performance in the vast majority of cases compared to existing approaches.

\end{abstract}

\ccsdesc[500]{Computing methodologies~Reasoning about belief and knowledge}

\keywords{Knowledge Graph; Knowledge Graph Embedding; Clustering; Continual Learning}

\maketitle

\section{Introduction}

\begin{figure}[t]
  \centering
  \includegraphics[width=0.9\columnwidth]{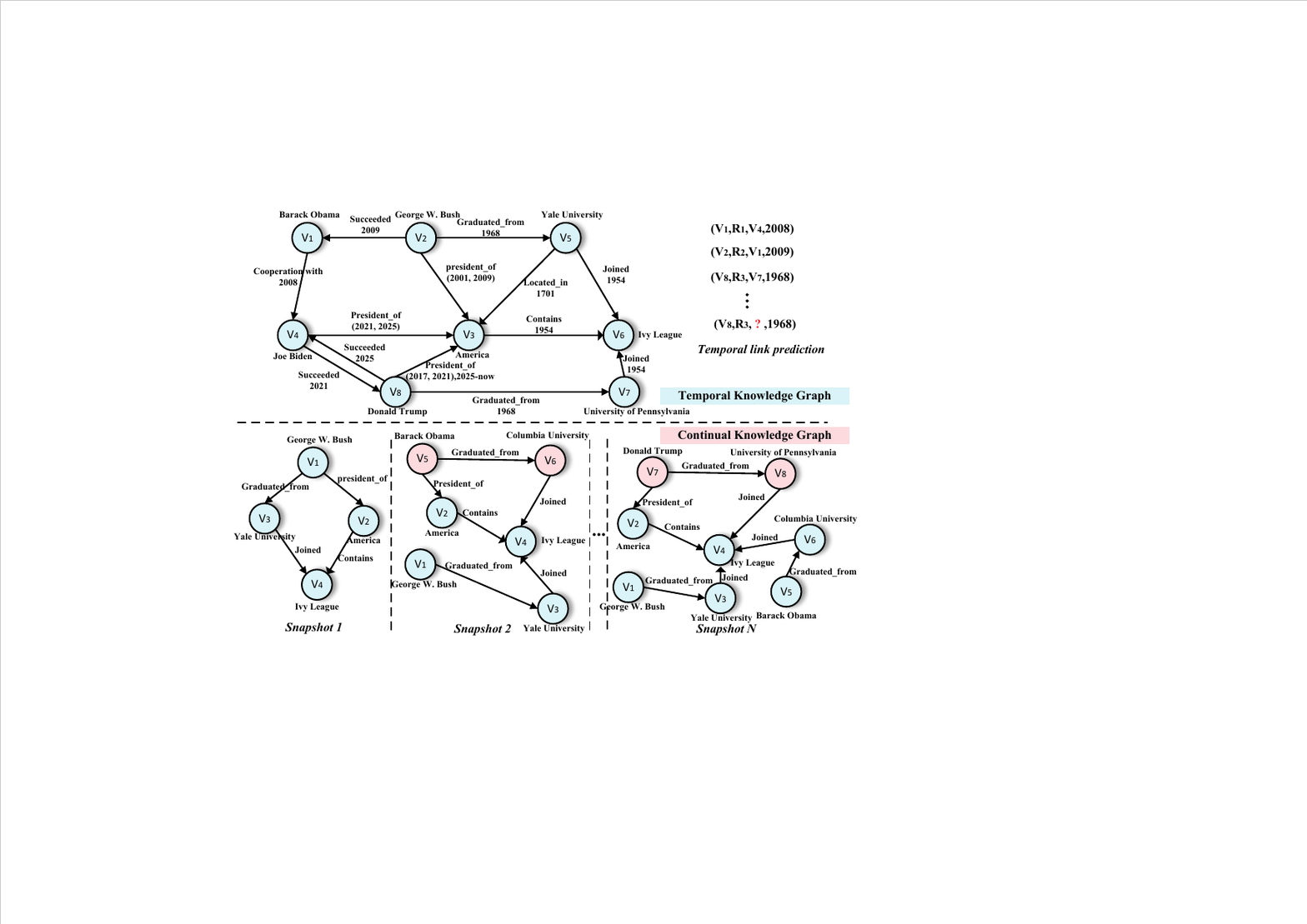}
  \caption{TKG uses a timestamp quadruple to explicitly annotate temporal information, while CKG decomposes the knowledge graph into time-unmarked evolutionary snapshots, focusing on the continuous updating and migration of knowledge.}
  \label{fig1}
\end{figure}
\renewcommand{\shortauthors}{Linyu Li, Zhi Jin et al.}

 Knowledge graphs (KGs)\cite{liang2024survey}\cite{ji2021survey} have become indispensable tools for representing structured information, supporting a wide range of knowledge‑driven applications such as LLM‑based question answering\cite{pan2024unifying}\cite{xu2025harnessing}, Blockchain\cite{song2025aero,song2022domain}, and recommender systems\cite{zhao2024leave}. Knowledge graph embedding (KGE) aims to encode entities and relations in a KG as continuous vectors, and the quality of these embeddings directly affects downstream performance. Classic KGE models \cite{bordes2013translating}\cite{sun2019rotate}\cite{li2025rethinking} perform excellently on static KGs, yet real‑world graphs usually evolve continuously. For example, the YAGO knowledge base evolved from YAGO3\cite{mahdisoltani2013yago3} to YAGO4.5\cite{suchanek2024yago}, with the number of entities soaring from about 4.6 million to 49 million and facts (triples) from 40 million to 132 million. For such continually evolving knowledge graphs (CKGs), existing KGE models must retrain the entire graph after each update—a computationally prohibitive and impractical strategy for large‑scale graphs like YAGO. Hence, this work focuses on CKGE models capable of learning CKGs efficiently.

 Recent CKGE research focuses on optimizing training structure, parameter efficiency, and regularization and masking mechanisms. In terms of structural optimization, IncDE \cite{liu2024towards} employs a layered strategy and incremental distillation mechanism to fully leverage graph structure information. In terms of parameter efficiency, FastKGE \cite{liu2024fast} reduces training parameters through incremental low-rank adapters, and ETT-CKGE \cite{zhu2025ett} uses task-driven tokens for efficient knowledge transfer. In terms of regularization and masking mechanisms, existing methods include lifelong learning based on masked autoencoders \cite{cui2023lifelong}, alignment of old and new knowledge based on energy models \cite{caocontinual}, biologically inspired dual masking mechanisms \cite{song2024orchestrating}, and flexible regularization based on the Fisher information matrix \cite{zhu2024flexible}. However, these methods inherently focus on passively "suppressing" forgetting through heuristic regularization terms or complex masking mechanisms, limiting the drastic changes in model parameters when learning new knowledge. While they mitigate catastrophic forgetting to some extent, they fail to provide a fundamental, principled framework for actively guiding knowledge evolution. In other words, they lack a foundational principle for clearly modeling how knowledge accumulates over time, instead focusing on preventing parameter corruption rather than guiding parameter evolution correctly.

As shown in Figure \ref{fig1}, CKGs differ from temporal KGs\cite{wang2023survey,ding2025halo}. While the latter model explicitly models temporal dependencies through timestamp relationships, CKGs focus on incremental updates to the graph structure itself, such as the emergence of new entities or relationships over time\cite{cui2023lifelong}. We keenly observe that due to the lack of direct temporal semantics in persistent knowledge graphs, their evolution is perfectly aligned with the sequential update mechanism of the Bayesian framework. This addresses the shortcomings of existing methods in actively guiding knowledge evolution. Following the Bayesian principle \cite{lee2024learning,bonnet2025bayesian}, we use the posterior distribution of the KG at snapshot $t$ as the prior for snapshot $t+1$, thereby gradually accumulating knowledge in a principled manner. This sequential update mechanism is insensitive to the order of knowledge and can effectively resist the forgetting of previous knowledge during the process of knowledge evolution. At the same time, it balances learning and forgetting through uncertainty quantification\cite{ahn2019uncertainty} (such as the probability distribution of weights), avoids catastrophic memory problems, and ensures that key knowledge is retained while outdated information is gradually released.

Furthermore, suppressing forgetting at the parameter level alone is insufficient. We identified another key challenge as semantic drift at the representation level: the evolution of knowledge representations between KG snapshots needs to be constrained, otherwise entity and relation embeddings in the latent space will drift unordered from snapshot to snapshot. This uncontrolled drift directly destroys the geometric structure of the embedding space, exacerbating catastrophic forgetting because the model cannot maintain the relative semantic relationships between entities, rendering distance- or angle-based scoring functions ineffective. To address this issue, we proposed a continual clustering method. This method acts as a semantic regularizer, constraining entity embeddings to maintain a compact cluster structure during their evolution through contrastive clustering. This ensures that entities remember not only who they are (intra-cluster compactness) but also who they are related to (inter-cluster separability), better ensuring that semantically similar entities maintain their relative positions in the embedding space over time, providing the model with clear guidance on how knowledge should be historically consistent. Our contributions are as follows:

\begin{itemize}
    \item  We propose the BAKE model, a novel Bayesian-guided CKGE model. This model treats CKGE as a sequential Bayesian inference problem, offering a novel approach to mitigating catastrophic forgetting.

    \item  We propose a continual clustering method for CKGE that constrains the evolution of knowledge between KG snapshots, maintaining semantic consistency while allowing controlled adaptation to new knowledge. This enables the model to remember not only the identity of entities (intra-cluster compactness) but also their relationships (inter-cluster separability).
    \item We conduct extensive experiments on eight CKGE datasets. Across various experimental results, the BAKE model significantly outperforms existing state-of-the-art methods in both knowledge preservation and adaptability.
\end{itemize}

\section{Preliminaries}

\subsection{Continual knowledge graph}
A Continual Knowledge Graph (CKG) is represented as a series of snapshots $\mathcal{G}_c = {\mathcal{S}_0, \mathcal{S}_1, \dots, \mathcal{S}_N}$, where each snapshot $\mathcal{S}_i = {\mathcal{E}_i, \mathcal{R}_i, \mathcal{T}_i}$ describes the state of the knowledge graph at time $i$. Each triple in $\mathcal{S}_i$ is represented as $(e_h, r, e_t) \in \mathcal{T}_i$, where $e_h$ and $e_t$ are the head and tail entities drawn from $\mathcal{E}_i$, and $r$ is a relation from the relation set $\mathcal{R}_i$. This study focuses on scenarios where relations remain constant over time (i.e., $\mathcal{R}_i = \mathcal{R}$ for all $i$), but entities expand gradually. The new entities and triples at time $i$ are denoted by $\Delta \mathcal{E}_i = \mathcal{E}_i - \mathcal{E}_{i-1}$ and $\Delta \mathcal{T}_i = \mathcal{T}_i - \mathcal{T}_{i-1}$, respectively. This setting aligns with typical real-world scenarios and is applied in current continual knowledge graph embedding methods.

\subsection{Continual Knowledge Graph Embedding}
The goal of Knowledge Graph Embedding is to embed entities and relations into a low-dimensional vector space while preserving their semantic meanings. For a CKG $\mathcal{G}_c = {\mathcal{S}_0, \mathcal{S}_1, \dots, \mathcal{S}_i}$, CKGE progressively builds an embedding model $\mathcal{M}_i$ as new snapshots $\mathcal{S}_i$ arrive. It leverages prior embeddings from snapshots ${\mathcal{S}_0, \dots, \mathcal{S}_{i-1}}$, optimizing training efficiency and overall performance. Completely retraining $\mathcal{M}_i$ from scratch on all triples in $\mathcal{S}_i$ is a basic yet inefficient choice unsuitable for practical scenarios. Current CKGE methods adapt the existing model $\mathcal{M}_{i-1}$ to new data $\mathcal{M}_i$ by creating embeddings for new entities $\Delta \mathcal{E}_i$ and refining embeddings for existing relations $\mathcal{R}$ and prior entities $\mathcal{E}_{i-1}$ when receiving new triples $\Delta \mathcal{T}_i$. CKGE is evaluated through continual link prediction: given an incomplete triple such as $(h,r,?)$, the model scores each candidate tail entity and outputs the highest-ranked one, maintaining accuracy while learning new knowledge and remembering past knowledge.

\subsection{Continual learning and Bayesian neural networks}
Continual learning (CL) aims to train a neural network on a sequence of tasks
$\mathcal{D}_1, \mathcal{D}_2, \dots, \mathcal{D}_T$
without forgetting previously acquired knowledge.
In standard CL, the model parameters $\theta$ are updated by stochastic gradient
descent, which often overwrites older information.

Bayesian neural networks (BNNs) \cite{bonnet2025bayesian}
provide a probabilistic framework for CL by treating the parameters $\theta$ as
random variables with a prior distribution $p(\theta)$.
The goal is to compute the posterior distribution $p(\theta \mid \mathcal{D})$
after observing data $\mathcal{D}$, according to Bayes’ rule:$p(\theta \mid \mathcal{D})=\frac{p(\mathcal{D} \mid \theta)\,p(\theta)}{p(\mathcal{D})}$.
Here, $p(\mathcal{D} \mid \theta)$ is the likelihood, and
$p(\mathcal{D})$ is the marginal likelihood.
For a task sequence, the posterior can be updated recursively as: $p(\theta \mid \mathcal{D}_{1:t})   \propto p(\mathcal{D}_t \mid \theta)\,p(\theta \mid \mathcal{D}_{1:t-1})$, meaning that the previous posterior serves as the new prior. In practice, we operationalize the Bayesian principle as a KL-based regularizer: on top of the standard KGE loss (Eq. 1), we add a precision-weighted constraint to the previous posterior (Eq. 4), which protects high-certainty knowledge while allowing controlled updates on low-certainty dimensions.

Exact inference is usually intractable for deep networks, so an independent Gaussian (mean‑field) approximation is commonly adopted \cite{lee2024learning,ahn2019uncertainty}: $q_t(\boldsymbol{\omega}) =\prod_{i=1}^{P} \mathcal{N}\!\bigl(\omega_i;\mu_{t,i},\sigma_{t,i}^2\bigr)$, with $\theta_t=(\boldsymbol{\mu}_t,\boldsymbol{\sigma}_t)$
taken as the learnable variational parameters. The independence assumption greatly simplifies the KL‐divergence computation
and the reparameterization gradient estimator, enabling iterative posterior updates.
In practice, the reparameterization trick $\omega_i=\mu_{t,i}+\sigma_{t,i}\,\epsilon,\quad \epsilon \sim \mathcal{N}(0,1)$ is often employed to estimate gradients.

\section{Methodology}

\subsection{Framework Overview}
\begin{figure*}[ht]
  \centering
  \includegraphics[width=0.9\textwidth]{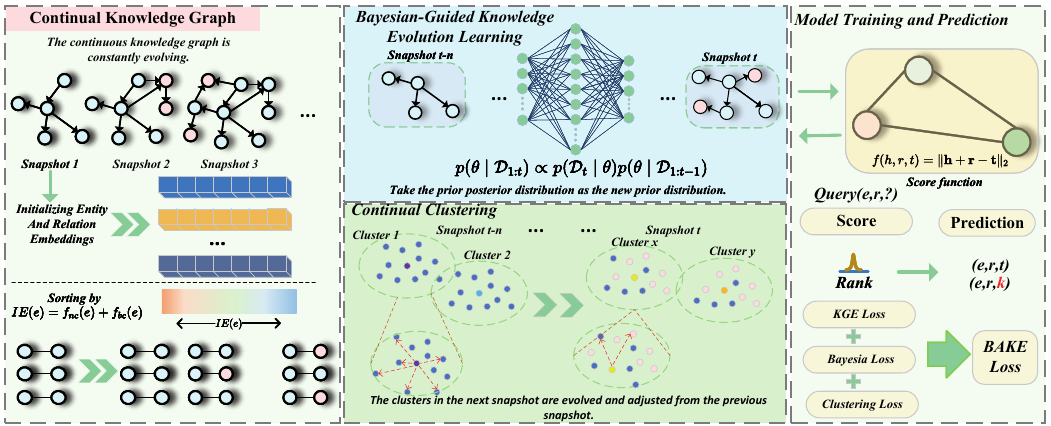}
  \caption{An overview of our proposed BAKE framework.}
  \label{fig2}
\end{figure*}

The overall framework of BAKE is shown in Figure \ref{fig2}. Its core idea is to formalize the continual knowledge graph embedding (CKGE) problem as a serialized Bayesian inference process. Given a knowledge graph sequence $\mathcal{G}_1, \mathcal{G}_2, \dots, \mathcal{G}_T$ that evolves over time, the model receives a set of newly added triples $\Delta \mathcal{T}_t$ at each time snapshot $t$. BAKE works together through the following three modules: First, Bayesian-guided knowledge evolution uses the parameter posterior distribution of the previous snapshot $t-1$ as a prior, integrates new knowledge and quantifies uncertainty through Bayesian updating; second, the continual clustering module introduces fairness regularization to maintain the semantic consistency of the embedding space; finally, the joint optimization objective integrates the above modules to achieve a balance between new knowledge learning and old knowledge retention. This framework aims to continually accumulate knowledge and alleviate catastrophic forgetting guided by the Bayesian rule.

\subsection{Bayesian‐Guided Knowledge Evolution Learning}
\label{sec:bayesian}
Inspired by related work such as uncertainty regularized continual learning \cite{ahn2019uncertainty}\cite{lee2024learning}\cite{bonnet2025bayesian}, we regard the embedding parameters of entities and relations as random variables in probability distributions rather than fixed point estimates. This approach not only facilitates quantification of parameter uncertainty but also protects important old knowledge when learning new knowledge, avoiding catastrophic forgetting.

\textbf{Entity and Relation Distributional Representations.} For each entity $e$ and relation $r$, their embedding vectors at time snapshot $t$ are modeled as Gaussian distributions. The embedding of entity $e$ is represented as $\mathbf{e}_t \sim \mathcal{N}(\boldsymbol{\mu}_{e,t}, (\text{diag}(\boldsymbol{\lambda}_{e,t}))^{-1})$, where $\boldsymbol{\mu}_{e,t} \in \mathbb{R}^d$ is the mean vector, and $\boldsymbol{\lambda}_{e,t} \in \mathbb{R}^d$ is the diagonal precision vector (inverse of variance); the embedding of relation $r$ is similar, represented as $\mathbf{r}_t \sim \mathcal{N}(\boldsymbol{\mu}_{r,t}, (\text{diag}(\boldsymbol{\lambda}_{r,t}))^{-1})$. Each element of the precision $\boldsymbol{\lambda}$ reflects the certainty of the corresponding dimension: a higher value indicates more reliable knowledge in that dimension, which should be better protected in subsequent updates.

\textbf{Sequential Bayesian Update.} At time snapshot $t$, we use the posterior distribution $\{\boldsymbol{\mu}_{t-1}, \boldsymbol{\lambda}_{t-1}\}$ from time $t-1$ as the prior and perform online Bayesian update using new data $\Delta \mathcal{T}_t$. To simplify inference, the embeddings $\hat{\boldsymbol{\Theta}}_t$ obtained from training on the current snapshot are treated as new observations to the prior mean $\boldsymbol{\mu}_{t-1}$, generated by the classic KGE model TransE~\cite{bordes2013translating}. TransE evaluates the plausibility of triples through a scoring function and optimizes the embeddings $\hat{\boldsymbol{\Theta}}_t=\{\hat{\mathbf{e}}_t,\,\hat{\mathbf{r}}_t\}$ by minimizing the margin-based ranking loss $\mathcal{L}_{\text{KGE}}$:
\begin{equation}
\small
\mathcal{L}_{\mathrm{KGE}}=\sum_{(h, r, t) \in \mathcal{T}} \sum_{\left(h^{\prime}, r, t^{\prime}\right) \in \mathcal{T}^{-}}\left[\gamma+f(h, r, t)-f\left(h^{\prime}, r, t^{\prime}\right)\right],
\end{equation}
where $f(h,r,t)=\lVert\mathbf{h}+\mathbf{r}-\mathbf{t}\rVert_2$ represents the score function of TransE \cite{bordes2013translating}. Based on the conjugate property of Gaussian distributions, the update rules for the embedding parameters of entity $e$ are:
\begin{align}
\boldsymbol{\lambda}_{e,t} &= \boldsymbol{\lambda}_{e,t-1} + \lambda_{obs}, \label{eq:prec_update} \\
\boldsymbol{\mu}_{e,t} &= \frac{\boldsymbol{\lambda}_{e,t-1} \odot \boldsymbol{\mu}_{e,t-1} + \lambda_{obs} \odot \hat{\mathbf{e}}_t}{\boldsymbol{\lambda}_{e,t}},
\end{align}
where $\hat{\mathbf{e}}_t$ is the entity embedding trained by TransE, $\lambda_{obs}$ is the fixed observation precision that controls the influence of new knowledge, and $\odot$ denotes element-wise multiplication. New entities are initialized with uninformative priors (random means, small precisions), and relation embeddings are updated in the same way. In experiments, we treat the value of $\lambda_{obs}$ as a tunable hyperparameter of the model to balance the importance of new and old knowledge.

To guide the embeddings towards the posterior mean, we use a regularization term \cite{lee2024learning} $\mathcal{L}_{Bayes}$, which is equivalent to minimizing the KL divergence between the current distribution $q(\mathbf{\Theta}_t)$ and the target posterior:
\begin{equation}
\mathcal{L}_{Bayes} = \sum_{i \in \mathcal{E}_t \cup \mathcal{R}_t} \beta \cdot \left\| \sqrt{\boldsymbol{\lambda}_{i,t-1}} \odot (\hat{\boldsymbol{\theta}}_{i,t} - \boldsymbol{\mu}_{i,t-1}) \right\|_2^2,
\label{eq:bayes_loss}
\end{equation}
where $\hat{\boldsymbol{\theta}}_{i,t}$ is the current embedding, and $\beta$ is the regularization hyperparameter. This term is weighted by the precision $\boldsymbol{\lambda}_{t-1}$, limiting changes to highly certain knowledge, thereby preserving old knowledge.


\subsection{Continual Clustering}
Bayesian updates prevent forgetting at the parameter level, but the evolution of knowledge representations between KG snapshots needs to be constrained; otherwise, entity and relation embeddings in the latent space may undergo uncertainty drift across snapshots. Inspired by contrastive learning and clustering methods~\cite{liang2024clustering}\cite{truong2025falcon}\cite{nguyen2021clusformer}, we propose a continual clustering method as a constraint to maintain the geometric structural consistency of the embedding space at the semantic level.

\textbf{Sequential Contrastive Clustering Learning.} In continual learning, imbalances in data volume and class distribution across temporal snapshots may lead to under-learning of minority classes. To address this, we propose the sequential contrastive clustering method ($\mathcal{L}_{FCC}$), which achieves fairness by dynamically adjusting the contributions of major and minor classes. Specifically, for each cluster $k$, we maintain a centroid vector $\mathbf{c}_k \in \mathbb{R}^d$ (where $d$ is the embedding dimension) to represent the central features of the cluster. First, we acknowledge the perspective proposed in~\cite{liu2024towards} that the learning order is also crucial in the process of learning KG structures. Second, since new entities emerge in each snapshot, making it difficult to apply k-means clustering, we sort the entities based on their importance, then fix the size of each cluster $k$, and take the mean of the embeddings in the current cluster $k$ as the clustering centroid $\mathbf{c}_k$. Specifically, the sorting process is as follows: we comprehensively consider the node's centrality and betweenness centrality in the graph structure to compute its importance score. The calculation formulas for these two metrics are as follows:
\begin{equation}
f_{nc}(e) = \frac{f_{\text{neighbor}}(e)}{N-1}, \quad f_{bc}(e) = \sum_{s,t \in \mathcal{E}, s \neq t} \frac{\sigma(s, t | e)}{\sigma(s, t)}
\end{equation}
where $f_{nc}(e)$ reflects the tightness of local connections for the entity, and $f_{bc}(e)$ measures its bridging role in information propagation. The variables in the formulas are defined as follows: $f_{\text{neighbor}}(e)$ represents the number of neighbors of entity $e$, $N$ is the total number of entities in the current knowledge graph snapshot, $\mathcal{E}$ is the set of entities in the current time snapshot $i$, $\sigma(s, t)$ is the total number of shortest paths between entities $s$ and $t$, and $\sigma(s, t \mid e)$ is the number of those paths that pass through entity $e$. We combine the two to define the importance score for entity $e$ as: $IE(e)=f_{nc}(e)+f_{bc}(e)$. We calculate the IE score for each entity and sort them in descending order of scores. Then, we assign the sorted entities to clusters of fixed size in sequence, and compute the initial embedding for each cluster as $\mathbf{c}_k=\frac{1}{|K|} \sum_{\mathbf{e}_i \in K} \mathbf{e}_i$. The continual clustering then constrains the representations of entities across different snapshots through contrastive loss, with the loss function defined as:

\begin{equation}
\small
\mathcal{L}_{\text{FCC}}=
-\sum_{k=1}^{K}\Bigl(
\alpha_k\!\!\sum_{\mathbf{e}_i\in\text{Cluster}_k}
\mathcal{L}_{\text{cont}}(\mathbf{e}_i,\mathbf{c}_k)+
\mathcal{L}_{\text{cont}}(\mathbf{v}_k,\mathbf{c}_k)
\Bigr)
\end{equation}
where $\mathbf{e}_i$ is the entity embedding vector belonging to cluster $k$, $\text{sim}(\cdot, \cdot)$ is the cosine similarity function, $\tau$ is the temperature parameter (controlling the smoothness of softmax), $K$ is the total number of clusters (equal to the number of entity classes in the current snapshot), and $\mathcal{L}_{\text{cont}}(\mathbf{e}_i, \mathbf{c}_k)$ is the standard contrastive loss, defined as:
\begin{equation}
\mathcal{L}_{\text{cont}}\left(\mathbf{e}_i, \mathbf{c}_k\right)=-\log \frac{\exp \left(\operatorname{sim}\left(\mathbf{e}_i, \mathbf{c}_k\right) / \tau\right)}{\sum_{j=1}^K \exp \left(\operatorname{sim}\left(\mathbf{e}_i, \mathbf{c}_j\right) / \tau\right)}.
\end{equation}
The scaling factor $\alpha_k$ is an adjustable parameter for each cluster, used to balance inter-cluster contributions (e.g., dynamically set as $\alpha_k = 1 / N_k$ based on the cluster sample size $N_k$ to boost the weight of minority classes). The learnable vector $\mathbf{v}_k \in \mathbb{R}^d$ serves as the proxy vector for cluster $k$, used to optimize the cluster center position, thereby balancing class contributions and promoting cluster compactness. This loss, by minimizing $\mathcal{L}_{FCC}$, pulls embeddings of entities in the same cluster closer to the centroid while pushing apart those from different clusters, achieving semantic consistency. Hyperparameters include the scaling factor $\alpha_k$, the temperature parameter $\tau$ , and the cluster sample size $N_k$.

\textbf{Cluster Maintenance and Update.}
At snapshot $t$, the centroids $\mathbf{c}_k$ of old classes are inherited and fixed from $t-1$, serving as anchors for old knowledge. New classes initialize centroids via feature means. During training, entities are dynamically assigned to the nearest cluster, and centroids are slowly adjusted through a momentum update:
\begin{equation}
\mathbf{c}_{k,t} = (1 - \eta) \mathbf{c}_{k,t-1} + \eta \cdot \text{mean}(\{\mathbf{e}_i \in \text{Cluster}_k\}),
\end{equation}
where $\eta$ is a hyperparameter balancing the stability and adaptability of cluster centroids.

\subsection{Final Training Objective} 
The optimal objective of BAKE at snapshot $t$ is the combined loss:
\begin{equation}
\mathcal{L}_{total} = \mathcal{L}_{KGE} +  \mathcal{L}_{Bayes} + \mathcal{L}_{FCC}.
\label{eq:final_loss}
\end{equation}

\section{Experiments}
\subsection{Experimental Setup}

\textbf{Datasets.} To thoroughly verify the effectiveness and stability of BAKE, we conducted extensive experiments on a wide variety of datasets, including: ENTITY~\cite{cui2023lifelong}, RELATION~\cite{cui2023lifelong}, HYBRID~\cite{cui2023lifelong}, GraphLower~\cite{liu2024towards}, GraphEqual~\cite{liu2024towards}, GraphHigher~\cite{liu2024towards}, FB-CKGE~\cite{liu2024fast}, and WN-CKGE~\cite{liu2024fast}. These eight datasets encompass different evolution patterns, each with unique characteristics. Detailed statistics of the datasets are listed in TABLE \ref{tab:1}. Each dataset contains 5 snapshots. 

\textbf{Implementation Details.} All experiments are implemented in PyTorch\cite{paszke2019pytorch} and conducted on 8 NVIDIA Tesla V100 GPUs. We tune all hyperparameters via grid search, including embedding dimensions for entities and relations $ D \in\{50, 100, 200, 300\}$, batch sizes $B \in\{ 256, 512, 1024\}$, and number of clusters $K$ $\in\{256, 512, 1024\}$, among others. We use Adam as the optimizer with learning rates selected from \{1e-5, 1e-4, 1e-3\}. To ensure fairness, all reported results are averaged over five runs.




\begin{table*}[t]
\centering
\caption{Performance comparison of various models on ENTITY, RELATION, FB-CKGE, and WN-CKGE datasets. The best results are in \textbf{bold}, and the second-best results are \underline{underlined}.}
\label{tab:2}
\resizebox{\textwidth}{!}{
\begin{tabular}{lcccccccccccccccc}
\toprule
 & \multicolumn{4}{c}{\textbf{ENTITY}} & \multicolumn{4}{c}{\textbf{RELATION}} & \multicolumn{4}{c}{\textbf{FB-CKGE}} & \multicolumn{4}{c}{\textbf{WN-CKGE}} \\
\cmidrule(lr){2-5} \cmidrule(lr){6-9} \cmidrule(lr){10-13} \cmidrule(lr){14-17}
\multirow{-2}{*}{\textbf{Model}} & \textbf{MRR} & \textbf{H@1} & \textbf{H@3} & \textbf{H@10} & \textbf{MRR} & \textbf{H@1} & \textbf{H@3} & \textbf{H@10} & \textbf{MRR} & \textbf{H@1} & \textbf{H@3} & \textbf{H@10} & \textbf{MRR} & \textbf{H@1} & \textbf{H@3} & \textbf{H@10} \\
\midrule
PNN & 0.229 & 0.130 & 0.265 & 0.425 & 0.167 & 0.096 & 0.191 & 0.305 & 0.215 & 0.122 & 0.245 & 0.403 & 0.134 & 0.002 & 0.241 & 0.342 \\
CWR & 0.088 & 0.028 & 0.114 & 0.202 & 0.021 & 0.010 & 0.024 & 0.043 & 0.075 & 0.011 & 0.105 & 0.192 & 0.005 & 0.002 & 0.007 & 0.012 \\
GEM & 0.165 & 0.085 & 0.188 & 0.321 & 0.093 & 0.040 & 0.106 & 0.196 & 0.188 & 0.103 & 0.212 & 0.359 & 0.119 & 0.002 & 0.215 & 0.297 \\
EMR & 0.171 & 0.090 & 0.195 & 0.330 & 0.111 & 0.052 & 0.126 & 0.225 & 0.180 & 0.097 & 0.204 & 0.346 & 0.114 & 0.002 & 0.205 & 0.286 \\
DiCGRL & 0.107 & 0.057 & 0.110 & 0.211 & 0.133 & 0.079 & 0.147 & 0.241 & 0.149 & 0.091 & 0.160 & 0.261 & 0.057 & 0.001 & 0.155 & 0.166 \\
SI & 0.154 & 0.072 & 0.179 & 0.311 & 0.113 & 0.055 & 0.131 & 0.224 & 0.187 & 0.102 & 0.211 & 0.359 & 0.115 & 0.001 & 0.209 & 0.289 \\
EWC & 0.229 & 0.130 & 0.264 & 0.423 & 0.165 & 0.093 & 0.190 & 0.306 & 0.218 & 0.124 & 0.247 & 0.410 & 0.136 & 0.003 & 0.248 & 0.338 \\
LKGE & 0.234 & 0.136 & 0.269 & 0.425 & 0.192 & 0.106 & 0.219 & 0.366 & 0.208 & 0.113 & 0.238 & 0.403 & 0.144 & 0.007 & 0.259 & 0.347 \\
FastKGE & 0.239 & 0.146 & 0.271 & 0.427 & 0.185 & 0.107 & 0.213 & 0.359 & 0.223 & 0.131 & 0.257 & 0.405 & \underline{0.159} & \underline{0.015} & \underline{0.287} & 0.356 \\
IncDE & 0.253 & 0.151 & \underline{0.291} & 0.448 & 0.199 & 0.110 & 0.221 & 0.368 & 0.232 & 0.134 & \underline{0.271} & \underline{0.428} & 0.150 & 0.003 & 0.278 & 0.366 \\
ETT-CKGE & 0.260 & 0.158 & - & 0.456 & 0.200 & 0.112 & - & 0.369 & \underline{0.236} & \underline{0.137} & - & 0.428 & 0.153 & 0.008 & - & \underline{0.385} \\
CLKGE & 0.248 & 0.144 & 0.278 & 0.436 & 0.203 & 0.115 & \underline{0.226} & 0.379 & - & - & - & - & - & - & - & - \\
SAGE & \underline{0.280} & \underline{0.176} & - & \underline{0.477} & \underline{0.217} & \underline{0.122} & - & \textbf{0.397} & - & - & - & - & - & - & - & - \\
\midrule
\rowcolor{gray!12}
BAKE & \textbf{0.287} & \textbf{0.179} & \textbf{0.326} & \textbf{0.484} & \textbf{0.224} & \textbf{0.126} & \textbf{0.239} & \underline{0.387} & \textbf{0.257} & \textbf{0.157} & \textbf{0.299} & \textbf{0.451} & \textbf{0.172} & \textbf{0.019} & \textbf{0.295} & \textbf{0.404} \\
\rowcolor{gray!15}
Improve & \textit{2.5\%} & \textit{1.7\%} & \textit{12.0\%} & \textit{1.5\%} & \textit{3.2\%} & \textit{3.3\%} & \textit{5.8\%} & \textit{-} & \textit{8.9\%} & \textit{14.6\%} & \textit{10.3\%} & \textit{5.4\%} & \textit{8.2\%} & \textit{26.7\%} & \textit{2.8\%} & \textit{4.9\%} \\
\bottomrule
\end{tabular}
}
\end{table*}


\begin{table*}[ht]
\centering
\caption{Performance comparison of various models on GraphLower, GraphEqual, GraphHigher, and HYBRID datasets.}
\label{tab:3}
\begin{tabular}{lcccccccccccc}
\toprule
 & \multicolumn{3}{c}{\textbf{GraphLower}} & \multicolumn{3}{c}{\textbf{GraphEqual}} & \multicolumn{3}{c}{\textbf{GraphHigher}} & \multicolumn{3}{c}{\textbf{HYBRID}} \\
\cmidrule(lr){2-4} \cmidrule(lr){5-7} \cmidrule(lr){8-10} \cmidrule(lr){11-13}
\multirow{-2}{*}{\textbf{Model}} & \textbf{MRR} & \textbf{H@1} & \textbf{H@10} & \textbf{MRR} & \textbf{H@1} & \textbf{H@10} & \textbf{MRR} & \textbf{H@1} & \textbf{H@10} & \textbf{MRR} & \textbf{H@1} & \textbf{H@10} \\
\midrule
Fine-tune & 0.185 & 0.098 & 0.363 & 0.183 & 0.096 & 0.358 & 0.198 & 0.108 & 0.375 & 0.135 & 0.069 & 0.262 \\
PNN       & 0.213 & 0.119 & 0.407 & 0.212 & 0.118 & 0.405 & 0.186 & 0.097 & 0.364 & 0.185 & 0.101 & 0.349 \\
CWR       & 0.032 & 0.005 & 0.080 & 0.122 & 0.041 & 0.277 & 0.189 & 0.096 & 0.374 & 0.037 & 0.015 & 0.077 \\
GEM       & 0.170 & 0.084 & 0.346 & 0.189 & 0.099 & 0.372 & 0.197 & 0.109 & 0.372 & 0.136 & 0.070 & 0.263 \\
EMR       & 0.188 & 0.101 & 0.362 & 0.185 & 0.099 & 0.359 & 0.202 & 0.113 & 0.379 & 0.141 & 0.073 & 0.267 \\
DiCGRL    & 0.102 & 0.039 & 0.222 & 0.104 & 0.040 & 0.226 & 0.116 & 0.041 & 0.242 & 0.149 & 0.083 & 0.277 \\
SI        & 0.186 & 0.099 & 0.366 & 0.179 & 0.092 & 0.353 & 0.190 & 0.099 & 0.371 & 0.111 & 0.049 & 0.229 \\
EWC       & 0.210 & 0.116 & 0.405 & 0.207 & 0.113 & 0.400 & 0.198 & 0.106 & 0.385 & 0.186 & 0.102 & 0.350 \\
LKGE      & 0.210 & 0.116 & 0.403 & 0.214 & 0.118 & 0.407 & 0.207 & 0.120 & 0.382 & 0.207 & 0.121 & 0.379 \\
IncDE     & 0.228 & 0.129 & 0.426 & 0.234 & 0.134 & 0.432 & 0.227 & 0.132 & 0.412 & 0.224 & 0.131 & \underline{0.401} \\
SAGE      & \underline{0.237} & \underline{0.138} & \underline{0.432} & \underline{0.247} & \underline{0.147} & \underline{0.446} & \underline{0.239} & \underline{0.143} & \underline{0.426} & \underline{0.224} & \underline{0.131} & 0.400 \\
\midrule
\rowcolor{gray!15}
BAKE      & \textbf{0.247} & \textbf{0.147} & \textbf{0.448} & \textbf{0.257} & \textbf{0.154} & \textbf{0.463} & \textbf{0.241} & \textbf{0.145} & \textbf{0.439} & \textbf{0.228} & \textbf{0.146} & \textbf{0.419} \\
\rowcolor{gray!15}
Improve & \textit{4.2\%} & \textit{6.5\%} & \textit{3.7\%} & \textit{4.0\%} & \textit{4.8\%} & \textit{3.8\%} & \textit{0.8\%} & \textit{1.4\%} & \textit{3.0\%} & \textit{1.8\%} & \textit{11.5\%} & \textit{4.5\%} \\
\bottomrule
\end{tabular}
\end{table*}

\begin{table*}[ht]
\centering
\caption{Ablation study results on \textbf{ENTITY}, \textbf{RELATION}, \textbf{FB-CKGE}, and \textbf{WN-CKGE} datasets.}
\label{tab:4}
\resizebox{\textwidth}{!}{
\begin{tabular}{lcccccccccccccccc}
\toprule
 & \multicolumn{4}{c}{\textbf{ENTITY}} & \multicolumn{4}{c}{\textbf{RELATION}} & \multicolumn{4}{c}{\textbf{FB-CKGE}} & \multicolumn{4}{c}{\textbf{WN-CKGE}} \\
\cmidrule(lr){2-5} \cmidrule(lr){6-9} \cmidrule(lr){10-13} \cmidrule(lr){14-17}
\multirow{-2}{*}{\textbf{Model}} & \textbf{MRR} & \textbf{H@1} & \textbf{H@3} & \textbf{H@10} & \textbf{MRR} & \textbf{H@1} & \textbf{H@3} & \textbf{H@10} & \textbf{MRR} & \textbf{H@1} & \textbf{H@3} & \textbf{H@10} & \textbf{MRR} & \textbf{H@1} & \textbf{H@3} & \textbf{H@10} \\
\midrule
\textbf{BAKE}        & 0.279 & 0.172 & 0.326 & 0.484 & 0.207 & 0.126 & 0.239 & 0.377 & 0.257 & 0.157 & 0.299 & 0.451 & 0.172 & 0.019 & 0.295 & 0.404 \\
\textbf{w/o Bayesian}  & 0.259 & 0.165 & 0.301 & 0.458 & 0.196 & 0.119 & 0.223 & 0.354 & 0.236 & 0.151 & 0.279 & 0.422 & 0.159 & 0.018 & 0.272 & 0.382 \\
\textbf{w/o cluster}   & 0.264 & 0.153 & 0.313 & 0.476 & 0.197 & 0.113 & 0.228 & 0.368 & 0.240 & 0.137 & 0.287 & 0.446 & 0.161 & 0.017 & 0.280 & 0.398 \\
\bottomrule
\end{tabular}
}
\end{table*}

\textbf{Baselines.} To fully demonstrate the effectiveness of BAKE, we compare it with all recent CKGE models, including Fine-tune~\cite{cui2023lifelong} (based on fine-tuning with new triples), IncDE~\cite{liu2024towards} and FastKGE~\cite{liu2024fast} (based on incremental distillation and LoRA fine-tuning), LKGE~\cite{cui2023lifelong}, EWC~\cite{kirkpatrick2017overcoming}, SI~\cite{zenke2017continual}, DiCGRL~\cite{kou2020disentangle}, EMR~\cite{wang2019sentence}, CWR~\cite{lopez2017gradient}, PNN~\cite{rusu2016progressive}, FMR~\cite{zhu2024flexible}, as well as energy-based and fine-grained token-driven models ETT-CKGE~\cite{zhu2025ett}, SAGE\cite{li2025sage} and CLKGE~\cite{caocontinual}.

\textbf{Metrics.} We evaluate link prediction performance using MRR, Hits@1, Hits@3, and Hits@10, where higher is better. Following~\cite{bordes2013translating} for negative sampling, we rank candidate triples by replacing the head or tail entity. For each snapshot $i$, we average scores over the current and all previous test sets. Results are reported using the fully trained model.

\subsection{Results and Analysis}
\begin{figure*}[t]
  \centering
  \includegraphics[width=0.9\textwidth]{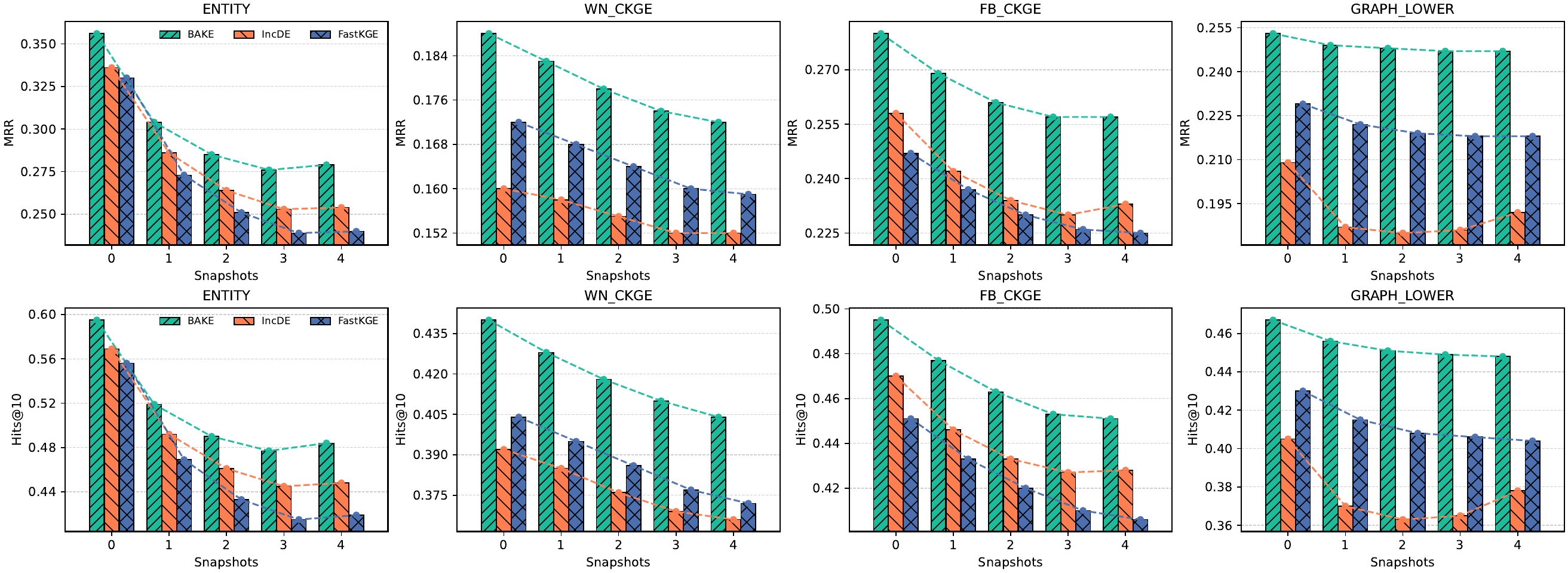}
  \caption{The change of model performance with the evolution of snapshots under ENTITY and FB\_CKGE datasets.}
    \label{fig:snapshots}
\end{figure*}

\begin{figure*}[t]
  \centering
  \includegraphics[width=0.9\textwidth]{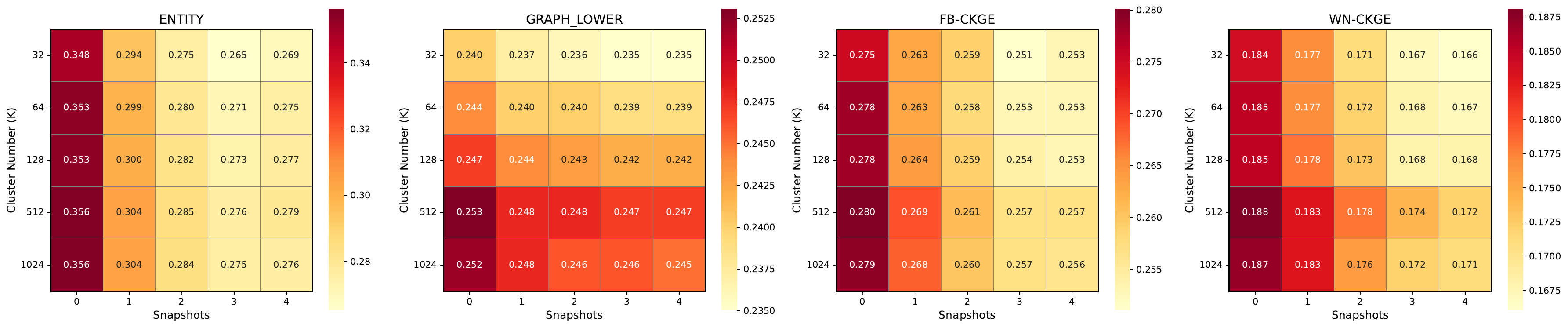}
  \caption{Parameter Sensitivity of BAKE to Cluster Number K under Different Snapshots on the ENTITY, GRAPH\_LOWER, FB\_CKGE, and WN\_CKGE Datasets.}
  \label{fig4}
\end{figure*}

\textbf{Main Results and Analysis.} 
Experimental results are shown in the Table \ref{tab:2} and Table \ref{tab:3}. Overall, our proposed BAKE method achieves state-of-the-art performance across all benchmark datasets and evaluation metrics (MRR, Hits@1/3/10), validating the effectiveness of the Bayesian-guided continual learning framework for the continual Knowledge Graph Embedding (CKGE) task. For large-scale initial knowledge graphs (e.g., FB-CKGE) and sparse or difficult scenarios (e.g., WN-CKGE), BAKE achieves state-of-the-art MRR and consistently maintains strong Hits, demonstrating robust knowledge acquisition while resisting catastrophic forgetting. These advantages are consistent with the design goal of combining sequential Bayesian updates with representation-level constraints.

Compared to representative baseline methods from different paradigms, including dynamic architecture methods (PNN, CWR), replay-based methods (GEM, EMR, DiCGRL), regularization-based methods (SI, EWC, LKGE), incremental/distillation or adapter-based methods (IncDE, FastKGE), and recent energy-based/token-driven methods (CLKGE, ETT-CKGE, SAGE), BAKE ranks first on the vast majority of datasets and metrics listed in the Tables \ref{tab:2} and \ref{tab:3}, and is at least competitive in the remaining cases. This trend holds across datasets with different growth dynamics: BAKE maintains optimal or near-optimal results for both balanced (GraphEqual) and non-balanced (GraphHigher/GraphLower) evolution modes, demonstrating that the proposed continual clustering constraint effectively stabilizes the embedding geometry under various evolution modes.

From a temporal perspective, BAKE also demonstrates the most stable performance across time snapshots: as the number of snapshots increases, its MRR and hits decay curves are consistently flatter than those of strong baselines such as IncDE/FastKGE, demonstrating its optimal balance between plasticity and stability. In summary, transforming the continual knowledge graph embedding problem into sequential Bayesian inference with explicit semantic consistency constraints provides a more principled and generally effective approach to existing strategies based on heuristic regularization, replay, or architectural expansion.

\textbf{Ablation Results and Analysis.} As shown in Table \ref{tab:4}, we conducted ablation experiments on four datasets, demonstrating that our proposed method significantly impacts performance. First, removing the Bayesian (i.e., directly using the IncDE distillation method) significantly degrades the overall model performance: MRR decreases by approximately 7.05\% on average, and Hits@10 decreases by approximately 5.84\% (e.g., MRR decreases by 8.17\% on FB-CKGE and H@10 decreases by 5.37\% on ENTITY). This demonstrates that sequential posterior updates effectively preserve prior knowledge and mitigate catastrophic forgetting caused by snapshot evolution. In contrast, removing Continual Clustering leads to steeper fluctuations in H@1: an average decrease of approximately 11.16\% (12.74\% and 10.32\% on FB-CKGE and RELATION, respectively), while H@10 only decreases slightly (1.66\% on average). This indicates that this mechanism primarily improves the strict first-place hit rate by maintaining semantic consistency and suppressing embedding drift. Overall, the collaboration of the two modules in BAKE enables it to achieve the optimal trade-off between adaptability and knowledge retention: the Bayesian is responsible for cross-snapshot steady-state memory, and continual clustering finely constrains local discrimination capabilities, thereby benefiting from MRR, Hits@10, and Hits@1 simultaneously.

\begin{figure*}[t]
  \centering
  \includegraphics[width=0.9\textwidth]{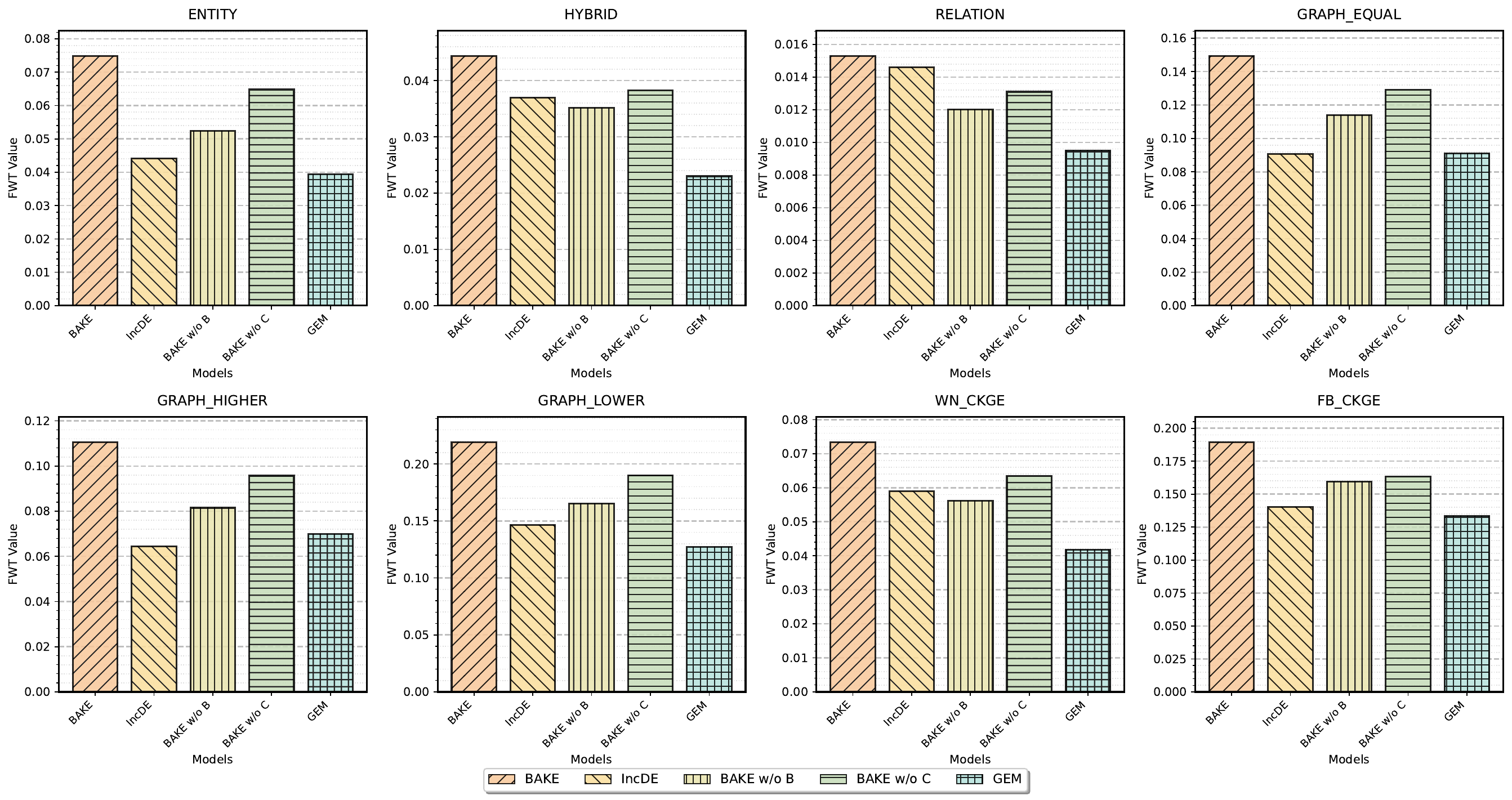}
  \caption{Knowledge transfer experiments on eight datasets. BAKE w/o B represents the use of knowledge distillation to retain the original knowledge without the guidance of Bayesian principles, and BAKE w/o C represents the use of no continual clustering method.}
  \label{fwt}
\end{figure*}

\begin{table}[htbp]

\centering
\caption{Case study experiments on the ENTITY dataset.}
\label{abtion}
\begin{tabular}{ll}
\toprule
\multicolumn{2}{c}{\textbf{Case 1}: ($Holland\ Taylor,\ profession,\ ?$)} \\
\midrule
\textbf{BAKE}     & Correct tail rank: ``\texttt{Actor}'' : \underline{1} \\
\textbf{IncDE}    & Correct tail rank: ``\texttt{Actor}'' : \underline{2} \\
\textbf{FastKGE}  & Correct tail rank: ``\texttt{Actor}'' : \underline{5} \\
\midrule
\multicolumn{2}{c}{\textbf{Case 2}: ($Blade\ Runner,\ genre,\ ?$)} \\
\midrule
\textbf{BAKE}     & Correct tail rank: ``\texttt{Drama}'' : \underline{1}  \\
\textbf{IncDE}    & Correct tail rank: ``\texttt{Drama}'' : \underline{26} \\
\textbf{FastKGE}  & Correct tail rank: ``\texttt{Drama}'' : \underline{3}  \\
\midrule
\multicolumn{2}{c}{\textbf{Case 3}: ($Netherlands,\ contains,\ ?$)} \\
\midrule
\textbf{BAKE}     & Correct tail rank: ``\texttt{Alkmaar}'' : \underline{1}  \\
\textbf{IncDE}    & Correct tail rank: ``\texttt{Alkmaar}'' : \underline{25} \\
\textbf{FastKGE}  & Correct tail rank: ``\texttt{Alkmaar}'' : \underline{69} \\
\midrule
\multicolumn{2}{c}{\textbf{Case 4}: ($Raoul\ Walsh,\ place\_of\_birth,\ ?$)} \\
\midrule
\textbf{BAKE}     & Correct tail rank: ``\texttt{New York}'' : \underline{3}   \\
\textbf{IncDE}    & Correct tail rank: ``\texttt{New York}'' : \underline{179} \\
\textbf{FastKGE}  & Correct tail rank: ``\texttt{New York}'' : \underline{336} \\
\bottomrule
\end{tabular}
\end{table}

\textbf{Results and analysis on different snapshots.} 
Figure ~\ref{fig:snapshots} shows the MRR trends of \textsc{BAKE} over five consecutive snapshots on the \textsc{ENTITY} and \textsc{FB-CKGE} datasets. Across all time steps, \textsc{BAKE} consistently outperforms strong baseline methods and exhibits the most stable performance trajectory: its MRR decreases more gradually with increasing snapshot index, and its fluctuations across snapshots are smaller. From a mechanistic perspective, the Bayesian approach's "posterior is prior" update strategy supports the rapid assimilation of new facts, while the proposed semantic clustering consistency regularization effectively suppresses embedding drift. These two components work together to achieve a favorable trade-off between plasticity and stability, rapidly adapting to new triples while minimizing the loss of existing knowledge.

\textbf{The parameter sensitivity experiment of cluster number K.} 
In a Continual learning scenario, using traditional K-means clustering independently at each snapshot can lead to cluster drift due to the dynamic emergence of entities and relations. To address this, we set K as a fixed hyperparameter and implement a continual clustering strategy across snapshots. As shown in the Figure \ref{fig4}, when K increases within a reasonable range, the MRR of BAKE improves overall, demonstrating that finer-grained clustering enhances semantic consistency and knowledge retention by providing a more accurate neighborhood structure. The performance differences between different K values are most pronounced in early snapshots and gradually decrease as the model evolves, indicating that the sensitivity to the number of clusters decreases after a stable representation is formed. However, excessively large K values can lead to semantic fragmentation and introduce additional overhead, resulting in diminishing returns or even negative effects. In practical applications, choosing a moderate K value can ensure robust performance while avoiding instability and unnecessary computational cost.

\textbf{Case Study.} To visually demonstrate BAKE's practical advantages, we conducted an in-depth analysis of four link prediction cases from the ENTITY dataset, comparing BAKE against strong baselines (IncDE and FastKGE). As shown in Table \ref{abtion}, these cases cover diverse relationship types, including profession, category, geographical affiliation, and birthplace. For queries such as \textit{(Blade Runner, genre, ?)} and \textit{(Netherlands, contains, ?)}, BAKE correctly ranked the ground-truth answers ("Drama" and "Alkmaar") at the top position. In contrast, competing models degraded the true answers to positions beyond the top 100. Remarkably, even for highly sparse relations like the query for Raoul Walsh's birthplace, BAKE achieved a competitive rank (3rd position), significantly outperforming the baselines (179th and 336th ranks).

\textbf{Knowledge Transfer Capability. } We evaluate knowledge transfer using the FWT metric on eight datasets. The experimental results are shown in Figure \ref{fwt}. BAKE attains the best FWT on all datasets, significantly outperforming IncDE (incremental distillation), GEM (replay-based), and both abridged variants. The gain is larger on datasets with more dramatic structural shifts (e.g., GraphHigher, GraphLower, FB-CKGE), showing that BAKE transfers learned concepts more effectively to later snapshots. Removing the Bayesian posterior update (BAKE w/o B) or the continual clustering constraint (BAKE w/o C) reduces FWT, indicating these components jointly suppress representation drift and improve forward transfer.

\section{Conclusion}
In this paper, we propose a novel Bayesian-guided CKGE model that effectively mitigates catastrophic forgetting in evolving knowledge graphs. Our framework comprises two complementary modules: (1) Bayesian-guided knowledge evolution, which treats sequential updates as Bayesian inference and uses posterior-as-prior updates to provide theoretical guarantees for knowledge preservation, and (2) continual clustering, which constrains semantic drift across snapshots. This ensures entities remember not only "who they are" (intra-cluster compactness) but also "who they are related to" (inter-cluster separability). Extensive experiments on multiple benchmarks demonstrate that our model significantly outperforms existing baselines in both knowledge preservation and adaptability, achieving optimal balance between plasticity and stability.
\begin{acks}
This work was supported by the Key Program of the National Natural Science Foundation of China (Grant No. 62436006).

\end{acks}

\bibliographystyle{ACM-Reference-Format}
\bibliography{sample-base}

\appendix


\section{Appendix}
\subsection{Detailed Experiment Settings}
\subsubsection{Datasets.}
\textbf{ENTITY~\cite{cui2023lifelong}:} This dataset simulates a growth scenario dominated by newly added entities. New entities are uniformly introduced and triple connections are supplemented at each time slice, requiring the model to learn new entity representations while preserving existing knowledge. This dataset contains five snapshots and serves as a benchmark for "new entity-driven evolution" in continual learning.

\textbf{RELATION~\cite{cui2023lifelong}:} This model depicts growth driven primarily by newly added relationships. New relationship types are added evenly over time, with the number of triples growing fastest. This model is used to test the stability of the model under frequent relationship additions. Five snapshots are also provided.

\textbf{HYBRID~\cite{cui2023lifelong}:} This model simulates "non-uniform dimensional growth," which is closer to reality. Entities, relationships, and triples all evolve, resulting in large incremental fluctuations and a higher level of challenge, which better tests the robustness of the method. Five snapshots are also provided.

\textbf{GraphEqual/Higher/Lower~\cite{liu2024towards}} all remove the restriction that new triples must contain old entities, to more closely resemble real-world graph evolution. They differ in their growth patterns: GraphEqual's increments are equal across all time periods, assessing continual learning under balanced growth; GraphHigher's increments increase gradually over time, simulating an accelerated burst of knowledge and testing the model's performance under sustained pressure; and GraphLower's increments are large early on and decreasing later, testing the model's ability to resist forgetting and consolidate knowledge during a "fast-first-then-slow" evolution.

\textbf{FB-CKGE~\cite{liu2024fast} and WN-CKGE~\cite{liu2024fast}} are two benchmarks built on FB15k-237 and WN18RR. They allocate about 60\% of triples in the initial snapshot, forming a large initial graph with a cold start, then gradually add the rest. This design tests methods for efficiency and anti-forgetfulness in large initial graphs with incremental updates.

\subsubsection{Baselines.}
 \textbf{Fin-tune (direct fine-tuning)~\cite{cui2023lifelong}} trains only on new triplets and often forgets old knowledge; \textbf{Elastic Weight Consolidation (EWC)} adds penalties on parameters important to past tasks to preserve them; \textbf{PNN (Progressive Neural Networks)~\cite{rusu2016progressive}} adds a new column per task with lateral connections to avoid forgetting, trading off with parameter growth.

\textbf{IncDE(Incremental Distillation)~\cite{liu2024towards} .} Uses hierarchical learning on new triplets with layer-wise distillation; a two-stage routine reduces early noise and better retains old knowledge.

\textbf{FastKGE (Incremental LoRA)~\cite{liu2024fast}.} Inserts LoRA adapters only where affected, adaptively sets ranks to cut parameters/training time, and uses layer selection to limit forgetting.

\textbf{LKGE (Lifelong KGE)~\cite{cui2023lifelong}} Updates embeddings via a masked KG autoencoder and uses transfer + regularization so new entities/relations inherit useful prior knowledge.

\textbf{SI (Synaptic Intelligence)~\cite{zenke2017continual}.} Online tracks each parameter’s contribution to loss reduction, assigns importance, and regularizes updates to protect critical weights.

\textbf{DiCGRL (Disentangle-based Continual Graph Representation Learning}~\cite{kou2020disentangle}). Separates transferable from task-specific factors in graph representations to learn new data while reducing forgetting.

\textbf{EMR (Episodic Memory Replay)~\cite{wang2019sentence}} replays a small buffer of old samples with new ones (EA-EMR aligns embeddings to reduce drift); \textbf{CWR (Copy Weights with Re-init)~\cite{lopez2017gradient}} resets a temporary head per task and copies it to a consolidated head to limit interference; \textbf{FMR (Flexible Memory Rotation)~\cite{zhu2024flexible}} rotates updates in parameter space (guided by Fisher information) and tunes regularization strength to balance stability and plasticity.

\textbf{ETT-CKGE (Efficient Task-driven Tokens for CKGE}~\cite{zhu2025ett}). Replaces explicit scoring/traversal with task-driven learnable tokens for alignment/transfer across snapshots, enabling simple matrix ops and lower training cost.

\textbf{SAGE (Scale-Aware Gradual Evolution}\cite{li2025sage}). Adapts embedding size to update scale and uses dynamic distillation to balance retention and absorption, yielding robust gains across benchmarks.

\textbf{CLKGE (Continual Learning Knowledge Graph Embeddings}~\cite{caocontinual}). Learns an energy manifold for dynamic KGs and aligns old/new energies to mitigate forgetting while transferring knowledge forward.

\subsubsection{FWT Metric.} To measure forward transferability, that is, the immediate help of the knowledge learned in the previous snapshot for the new task in the next snapshot, we use FWT to measure the ability of the model:
\begin{equation}
\mathrm{FWT}=\frac{1}{N} \sum_{i=1}^N a_{i-1, i},
\end{equation}
where $a_{t, i}$ represents the score of the model $M_t$ trained on the $t$th snapshot on the test set (often expressed as MRR or Hits@K) at the $i$th snapshot. Therefore, $a_{i-1, i}$ captures the model's zero-shot generalization ability to the next task before seeing $\Delta T_i$ training data. A larger FWT value indicates that the method is more effective in transferring learned structural and semantic patterns to subsequent snapshots.

\subsubsection{MRR Metric.} Mean Reciprocal Rank (MRR) evaluates the ranking quality by considering the reciprocal of the rank position of the correct answer. Formally, given a query set $\mathcal{Q}$ at snapshot $i$, let $\mathrm{rank}_q$ denote the rank of the ground-truth entity (or answer) for query $q \in \mathcal{Q}$. The MRR score is defined as:
\begin{equation}
\mathrm{MRR}=\frac{1}{|\mathcal{Q}|}\sum_{q\in\mathcal{Q}}\frac{1}{\mathrm{rank}_q}.
\end{equation}
Higher MRR indicates correct answers rank nearer the top.

\subsubsection{Hits@N Metric.} Hits@N (also denoted as Hits@K) measures the proportion of queries whose correct answers appear in the top-$N$ ranked results. Specifically, for each query $q \in \mathcal{Q}$, we define an indicator function $\mathbb{I}(\mathrm{rank}_q \le N)$, then Hits@N is computed as:
\begin{equation}
\mathrm{Hits@}N=\frac{1}{|\mathcal{Q}|}\sum_{q\in\mathcal{Q}}\mathbb{I}(\mathrm{rank}_q \le N).
\end{equation}
A larger Hits@N indicates that the model retrieves correct answers within the top-$N$ positions more frequently.

\begin{table*}[htbp]

\centering
\caption{The statistics of datasets. $N_E$, $N_R$ and $N_T$ denote the number of  entities and relations, and the current  snapshot $ i$}

\label{tab:1}
\begin{tabular}{l ccc ccc ccc ccc ccc} 
\toprule
& \multicolumn{3}{c}{Snapshot 0} & \multicolumn{3}{c}{Snapshot 1} & \multicolumn{3}{c}{Snapshot 2} & \multicolumn{3}{c}{Snapshot 3} & \multicolumn{3}{c}{Snapshot 4} \\
\cmidrule(lr){2-4} \cmidrule(lr){5-7} \cmidrule(lr){8-10} \cmidrule(lr){11-13} \cmidrule(lr){14-16}
Dataset & \multicolumn{1}{c}{$N_E$} & \multicolumn{1}{c}{$N_R$} & \multicolumn{1}{c}{$N_T$} & \multicolumn{1}{c}{$N_E$} & \multicolumn{1}{c}{$N_R$} & \multicolumn{1}{c}{$N_T$} & \multicolumn{1}{c}{$N_E$} & \multicolumn{1}{c}{$N_R$} & \multicolumn{1}{c}{$N_T$} & \multicolumn{1}{c}{$N_E$} & \multicolumn{1}{c}{$N_R$} & \multicolumn{1}{c}{$N_T$} & \multicolumn{1}{c}{$N_E$} & \multicolumn{1}{c}{$N_R$} & \multicolumn{1}{c}{$N_T$} \\
\midrule
ENTITY      & 2,909  & 233 & 46,388  & 5,817  & 236 & 72,111  & 8,275  & 236 & 73,785  & 11,633 & 237 & 70,506  & 14,541 & 237 & 47,326  \\
RELATION    & 11,560 & 48  & 98,819  & 13,343 & 96  & 93,535  & 13,754 & 143 & 66,136  & 14,387 & 190 & 30,032  & 14,541 & 237 & 21,594  \\
HYBRID      & 8,628  & 86  & 57,561  & 10,040 & 102 & 20,873  & 12,779 & 151 & 88,017  & 14,393 & 209 & 103,339 & 14,541 & 237 & 40,326  \\
GraphEqual  & 2,908  & 226 & 57,636  & 5,816  & 235 & 62,023  & 8,724  & 237 & 62,023  & 11,632 & 237 & 62,023  & 14,541 & 237 & 66,411  \\
GraphHigher & 900    & 197 & 10,000  & 1,838  & 221 & 20,000  & 3,714  & 234 & 40,000  & 7,467  & 237 & 80,000  & 14,541 & 237 & 160,116 \\
GraphLower  & 7,505  & 237 & 160,000 & 11,258 & 237 & 80,000  & 13,134 & 237 & 40,000  & 14,072 & 237 & 20,000  & 14,541 & 237 & 10,116  \\
FB-CKGE     & 7,505  & 237 & 186,070 & 11,258 & 237 & 31,012  & 13,134 & 237 & 31,012  & 14,072 & 237 & 31,012  & 14,541 & 237 & 31,010  \\
WN-CKGE     & 24,567 & 11  & 55,801  & 28,660 & 11  & 9,300   & 32,754 & 11  & 9,300   & 36,848 & 11  & 9,300   & 40,943 & 11  & 9,302   \\
\bottomrule
\end{tabular}

\end{table*}
\subsection{Related Work}
\subsubsection{Knowledge Graph Embedding}
Early KGE models mainly follow the translation paradigm, such as TransE\cite{bordes2013translating} and TransH\cite{wang2014knowledge} in the real vector space, and their extensions to complex/quaternion spaces, including RotatE\cite{sun2019rotate}, QuatE\cite{zhang2019quaternion}, HAKE\cite{zhang2020learning}, SATKGC\cite{ko2025subgraph}, MRME\cite{li2025multi}, SymAgent\cite{liu2025symagent}, HySAE\cite{li2025hysae}, and GIE\cite{cao2022geometry}. These approaches introduced the idea of representing a triple $(h,r,t)$ through simple algebraic operations in a vector space. Since then, both the efficiency and the expressive power of KGE models have improved along multiple dimensions. RecPiece\cite{liang2024clustering} clusters relation centres to select anchors, reducing memory consumption while improving link prediction accuracy; RAA‑KGC\cite{yuan2025knowledge} pulls queries of a pretrained language model toward relation‑aware neighbour anchors to construct more discriminative embeddings; FedE\cite{zhou2024poisoning} pioneers a poisoning attack on federated KG embedding, using inferred relations to craft stealthy updates that induce specific errors without degrading main task performance.
Recently, KGE research has begun to leverage diffusion models to recast the task as conditional entity generation. For example, FDM\cite{long2024fact} trains a diffusion model that outputs the distribution of tail entities conditioned on the head entity and relation, thereby capturing multimodal relational semantics without relying on complicated probabilistic mechanisms. S2DN\cite{ma2025s2dn} proposes a semantic-aware denoising network and filters out unreliable interactions to enhance the structural reliability of target links.

\subsubsection{Continual Knowledge Graph Embedding}
As knowledge graphs (KGs) evolve, embedding models must learn new facts while avoiding catastrophic forgetting of previously acquired knowledge. Research on continual knowledge graph embedding (CKGE)\cite{li2024learning} has progressed notably in recent years. IncDE~\cite{liu2024towards} orders new triples hierarchically and employs layer-wise incremental distillation to preserve old knowledge; FastKGE~\cite{liu2024fast} accelerates updates with an adaptive low-rank adapter (IncLoRA), substantially shortening training time while isolating parameters for new snapshots; LKGE~\cite{cui2023lifelong} combines a masked autoencoder with transfer regularization to extend embeddings across multiple snapshots; ETT-CKGE~\cite{zhu2025ett} replaces costly node scoring with lightweight task-driven labels and aligns KG snapshots through simple matrix operations; CLKGE~\cite{caocontinual} couples continual learning with an energy-based manifold so that new and old knowledge reinforce each other; CMKGE~\cite{song2024orchestrating} introduces a biologically inspired dual inhibition–excitation mask to balance plasticity and stability; FMR~\cite{zhu2024flexible} rotates the parameter space based on Fisher information and assigns flexible regularization to further mitigate forgetting; ERPP~\cite{yang2025knowledge} transfers relational paths across snapshots to shift from catastrophic forgetting to knowledge accumulation; CFKGC~\cite{li2024learning} introduces  model-level modulation to mitigate catastrophic forgetting while leveraging multi-view relation augmentation to better learn scarce emerging relations; and SAGE~\cite{li2025sage} adaptively expands embedding dimensions and uses dynamic distillation for CKGE.

\subsubsection{Continual Learning}
Continual learning (CL)\cite{wang2024comprehensive} enables models to learn from a continuous data stream, addressing the "catastrophic forgetting" of old knowledge, a problem first identified in neural networks by McCloskey and Cohen\cite{mccloskey1989catastrophic}. Unlike time series prediction \cite{zhang2025adamixt}, the goal is to balance stability (retaining knowledge) with plasticity.

CL strategies are now applied across various dynamic environments. Recent examples include Helios \cite{shi2025helios} for adapting to new cyber threats, DSLR \cite{choi2024dslr} for enhancing diversity in graph continual learning, and Multimodal Graph Continual Learning \cite{cai2022multimodal} over evolving document corpora. In recommender systems, Continual Collaborative Distillation \cite{lee2024continual} allows models to adapt to non-stationary user and item data. These advancements demonstrate CL's growing importance in web systems.

\subsection{BAKE model training process pseudocode}
To help readers more clearly and comprehensively understand our proposed BAKE model, we present its training process in detail using pseudocode. The core idea of BAKE is to formulate the CKGE task as a sequential Bayesian inference problem, thus providing a more theoretically complete framework for the continuous accumulation of knowledge and the suppression of forgetting.
The algorithm iterates through each knowledge graph snapshot sequentially. For a given snapshot $t$, the model utilizes the parameterized posterior distribution $(\boldsymbol{\mu}_{t-1}, \boldsymbol{\lambda}_{t-1})$ of the previous snapshot $t-1$ as a guiding prior. During training, the model jointly optimizes three objectives: a standard KGE loss (for learning new knowledge), a Bayesian regularization term guided by the prior (for preventing catastrophic forgetting of old knowledge), and a persistent clustering loss (for maintaining semantic consistency in the embedding space and preventing representation drift). After training on the current snapshot data, the model performs a Bayesian update to compute a new posterior distribution $(\boldsymbol{\mu}_t, \boldsymbol{\lambda}_t)$, which serves as the prior for the next snapshot $t+1$. Algorithm~\ref{alg:bake} describes in detail the complete steps from initialization to the final sequential update.


\subsection{Additional Discussion}

\subsubsection{End-user tasks and application relevance.}
Knowledge graph embeddings are a foundational representation for many knowledge-driven applications, including LLM-based question answering and recommender systems, where embedding quality directly affects downstream performance \cite{yuan2025hek}. In continual settings, these applications require models to rapidly absorb newly emerging facts while maintaining previously learned semantics. In this work, we follow the standard CKGE protocol by evaluating models on continual link prediction, i.e., ranking candidate entities for incomplete queries such as $(h,r,?)$ while learning new knowledge without forgetting past knowledge. This evaluation is not only widely adopted in prior CKGE studies but also closely aligned with real-world needs because link prediction is the core operation behind KG completion and retrieval. 

\subsubsection{Runtime and computational efficiency.}
We acknowledge the importance of computational efficiency in continual learning and complement our accuracy evaluation with an efficiency discussion. BAKE introduces lightweight overhead on top of the backbone KGE training: the sequential Bayesian posterior-as-prior update is implemented in closed form with per-dimension operations, as shown by the element-wise update rules and the element-wise product operator. The continual clustering component maintains centroids via a momentum update, which is performed once per epoch in our training procedure. In our implementation, BAKE's training time per snapshot is within about $1.2\times$ of IncDE; the main extra cost comes from centroid maintenance and reassignment, while Bayesian updates are negligible compared with standard mini-batch KGE optimization. All experiments are conducted under the same hardware setting (8 NVIDIA Tesla V100 GPUs).

\subsubsection{Choice of TransE as the base scorer.}
We use TransE as the base scorer to generate snapshot-specific embedding observations, following the standard KGE training objective with $f(h,r,t)=\|h+r-t\|_2$~\cite{bordes2013translating}. This design ensures direct comparability to many recent CKGE baselines that adopt the same backbone, while allowing us to isolate the contribution of the proposed continual Bayesian framework. Importantly, BAKE does \emph{not} rely on TransE-specific assumptions: in our formulation, the embeddings $\hat{\Theta}_t$ obtained from the snapshot learner are treated as observations used for Bayesian updating, and the posterior update itself is model-agnostic as long as a snapshot learner can produce $\hat{\Theta}_t$. Therefore, BAKE can, in principle, be combined with more expressive KGE scorers (e.g., RotatE, QuatE)\cite{sun2019rotate,zhang2019quaternion}, while we intentionally keep the base scorer fixed in this paper to provide a controlled and fair comparison across continual-learning strategies.

\begin{algorithm}[H]
\caption{BAKE Model Training Process}
\label{alg:bake}
\begin{algorithmic}[1]
\Require KG snapshot sequence $\mathcal{G}_c=\{\mathcal{S}_0,\dots,\mathcal{S}_N\}$; hyperparameters: lr, epochs $E$, reg.\ $\beta$, obs.\ precision $\lambda_{obs}$, momentum $\eta$, clusters $K$, temp.\ $\tau$.
\State \textbf{Init for $\mathcal{S}_0$:} init entity/relation embeddings $\hat{\boldsymbol{\Theta}}_0$; set priors $\boldsymbol{\mu}_0\!\gets\!\hat{\boldsymbol{\Theta}}_0$, small $\boldsymbol{\lambda}_0$; compute $IE(e)$ and cluster into $K$; get centroids $\mathbf{c}_{k,0}$ and proxies $\mathbf{v}_{k,0}$.
\For{$t=0\to N$}
  \If{$t>0$}
    \State Inherit priors $\{\boldsymbol{\mu}_{t-1},\boldsymbol{\lambda}_{t-1}\}$; init new entities $\Delta\mathcal{E}_t$; set $\mathbf{c}_{k,t}\!\gets\!\mathbf{c}_{k,t-1}$, $\mathbf{v}_{k,t}\!\gets\!\mathbf{v}_{k,t-1}$.
  \EndIf
  \For{$epoch=1\to E$} \Comment{Train on $\mathcal{S}_t$}
    \For{each batch in $\mathcal{T}_t$}
      \State Compute KGE loss $\mathcal{L}_{\text{KGE}}$ via eq.~1.
      \State Compute Bayesian reg.\ $\mathcal{L}_{\text{Bayes}}$ via eq.~4 using $\{\boldsymbol{\mu}_{t-1},\boldsymbol{\lambda}_{t-1}\}$.
      \State Reassign entities to nearest clusters; compute $\mathcal{L}_{\text{FCC}}$ via eq.~6.
      \State $\mathcal{L}_{\text{total}}\gets\mathcal{L}_{\text{KGE}}+\mathcal{L}_{\text{Bayes}}+\mathcal{L}_{\text{FCC}}$; update $\hat{\boldsymbol{\Theta}}_t$, $\mathbf{v}_{k,t}$.
    \EndFor
    \State Update centroids $\mathbf{c}_{k,t}$ with momentum via eq.~8.
  \EndFor
  \State \textbf{Bayesian update for next snapshot:}
  \State $\boldsymbol{\lambda}_{t}\gets\boldsymbol{\lambda}_{t-1}+\lambda_{obs}$ \hfill (via eq.~2)
  \State $\boldsymbol{\mu}_{t}\gets\frac{\boldsymbol{\lambda}_{t-1}\odot\boldsymbol{\mu}_{t-1}+\lambda_{obs}\odot\hat{\boldsymbol{\Theta}}_t}{\boldsymbol{\lambda}_{t}}$ \hfill (via eq.~3)
\EndFor
\State \Return final embeddings $\hat{\boldsymbol{\Theta}}_N$.
\end{algorithmic}
\end{algorithm}

\end{document}